\documentclass{article} 
  \usepackage{iclr2018_conference,times}
  \iclrfinalcopy 
  \usepackage{hyperref}
  \usepackage{url}

\usepackage[utf8]{inputenc} 
\usepackage[T1]{fontenc}    
\usepackage{booktabs}       
\usepackage{amsfonts}       
\usepackage{nicefrac}       
\usepackage{microtype}      

\newcommand{\varmisusetask}{\textsc{VarMisuse}\xspace}
\newcommand{\nametaskname}{\textsc{VarNaming}\xspace}
\newcommand{\ourtitle}{Learning to Represent Programs with Graphs}

\usepackage{amssymb}       

\usepackage{multirow}
\usepackage{paralist}

\usepackage{algorithm}
\usepackage[noend]{algorithmic}

\usepackage{xspace}
\usepackage{color}
\usepackage{amsmath}
\usepackage{subcaption}
\usepackage{bm}

\definecolor{darkgreen}{rgb}{0,0.7,0}

\newcommand{\cf}{\hbox{\emph{cf.}}\xspace}

\newcommand{\eg}{\hbox{\emph{e.g.}}\xspace}
\newcommand{\ie}{\hbox{\emph{i.e.}}\xspace}

\newcommand{\etc}{\hbox{\emph{etc.}}\xspace}

\newcommand{\CSharp}{C\nolinebreak\hspace{-.05em}\raisebox{.6ex}{\footnotesize\bf \#}}

\usepackage{ifthen}
\newboolean{showcomments}
\setboolean{showcomments}{true} 
\ifthenelse{\boolean{showcomments}}{
  \newcommand{\nbc}[3]{
    {\colorbox{#3}{\bfseries\sffamily\scriptsize\textcolor{white}{#1}}}%
    {\textcolor{#3}{\sf\small$\blacktriangleright$\textit{#2}$\blacktriangleleft$}}}
  \newcommand{\todo}[1]{\nbc{TODO}{#1}{blue}\xspace}
}{
  \newcommand{\nbc}[3]{}
  \newcommand{\todo}[1]{}
}

\newcommand{\eat}[1]{}

\newcommand{\rSC}[1]{Sect.~\ref{#1}}
\newcommand{\rF}[1]{Fig.~\ref{#1}}

\usepackage{listings}
\newcommand{\id}[1]{\texttt{#1}}

\DeclareMathOperator*{\argmax}{arg\,max}

\newcommand{\vect}[1]{\ensuremath{\mathbf{r}}\xspace}
\newcommand{\tok}{\ensuremath{t}\xspace}
\newcommand{\tokenSeq}{\ensuremath{\mathcal{T}}\xspace}
\newcommand{\var}{\ensuremath{v}\xspace}
\newcommand{\varSet}{\ensuremath{\mathbb{V}}\xspace}
\newcommand{\varType}[1]{\ensuremath{\tau}(#1)\xspace}
\newcommand{\allVarType}[1]{\ensuremath{\tau^*(#1)}\xspace}

\newcommand{\varsInScope}[1]{\ensuremath{\varSet_{#1}}\xspace}

\newcommand{\dataFlowSym}{\ensuremath{\mathcal{D}}}

\newcommand{\dfgReadPred}[1]{\ensuremath{\dataFlowSym^R(#1)}}
\newcommand{\dfgWritePred}[1]{\ensuremath{\dataFlowSym^W(#1)}}

\newcommand{\typeEmbedSym}[0]{\ensuremath{\mathbf{r}}}
\newcommand{\typeEmbed}[1]{\ensuremath{\typeEmbedSym(#1)}}
\newcommand{\typeSetEmbed}[1]{\ensuremath{\typeEmbedSym^*(#1)}}

\newcommand{\localContextSym}[0]{\ensuremath{{\bm{c}}}}
\newcommand{\localContextRepr}[1]{\ensuremath{\localContextSym(#1)}}

\newcommand{\usageRepr}[2]{\ensuremath{\mathbf{u}(#1, #2)}}

\newcommand{\localmodel}{\textsc{Loc}\xspace}
\newcommand{\avglblmodel}{\textsc{AvgLBL}\xspace}
\newcommand{\avgbirnnmodel}{\ensuremath{\textsc{AvgBiRNN}}\xspace}
\newcommand{\graphmodel}{\ensuremath{\textsc{GGNN}}\xspace}

\newcommand{\namedplaceholder}[1]{\colorbox{blue}{\textcolor{white}{\id{~#1~}}}}
\newcommand{\placeholder}[1]{\colorbox{yellow}{\textcolor{black}{\id{#1}}}}

\usepackage{tikz}
\usetikzlibrary{arrows,decorations.pathmorphing,backgrounds,fit,positioning,automata,shapes,calc,shadows}
\tikzset{codegraph/.style={
                every node/.style={anchor=north, text depth=0.5ex, text
                  height=1.5ex, text centered, scale=.75, inner sep=2pt},
                SyntaxToken/.append style={rectangle, draw=black, fill=black!10, font=\tt},
                SyntaxNode/.append style={rounded rectangle, draw=blue},
         }
}

\makeatletter
\newenvironment{btHighlight}[1][]
{\begingroup\tikzset{bt@Highlight@par/.style={#1}}\begin{lrbox}{\@tempboxa}}
{\end{lrbox}\bt@HL@box[bt@Highlight@par]{\@tempboxa}\endgroup}

\newcommand\btHL[1][]{%
  \begin{btHighlight}[#1]\bgroup\aftergroup\bt@HL@endenv%
}

\def\bt@HL@endenv{%
  \end{btHighlight}%
  \egroup
}
\newcommand{\bt@HL@box}[2][]{%
  \tikz[#1]{%
    \pgfpathrectangle{\pgfpoint{1pt}{0pt}}{\pgfpoint{\wd #2}{\ht #2}}%
    \pgfusepath{use as bounding box}%
    \node[anchor=base west, fill=blue!10,outer sep=0pt,inner xsep=1pt, rounded corners=1pt, inner ysep=0pt, minimum height=\ht\strutbox,#1]{\strut\usebox{#2}};
  }%
}
\makeatother

\newcommand{\hlreflarge}[1]{\begin{btHighlight}[fill=white!10,draw=green,solid,line width=.1pt]{$#1$}\end{btHighlight}}
\newcommand{\hlref}[1]{$^\text{\tiny~\hlreflarge{#1}}$}

\newcommand{\hlplacehld}[2]{\begin{btHighlight}[fill=red!20,draw=red,solid,line width=.1pt]{#1$^{#2}$}\end{btHighlight}}

\newcommand{\hlvar}[2]{\begin{btHighlight}[fill=black!10,draw=black,solid,line width=.1pt]{#1}\end{btHighlight}$\,^{#2}$}
\newcommand{\explain}{$\triangleright$\xspace}

\newcommand{\graph}{\mathcal{G}}
\newcommand{\nodes}{\mathcal{V}}
\newcommand{\node}{v}
\newcommand{\nodefeats}{\boldsymbol{X}}
\newcommand{\nodefeat}{\boldsymbol{x}}
\newcommand{\edgelist}{\boldsymbol{\mathcal{E}}}
\newcommand{\edges}{\mathcal{E}}

\newcommand{\statesymbol}{\boldsymbol{h}}
\newcommand{\state}[1]{\ensuremath{\statesymbol^{(#1)}}}

\newcommand{\statePrime}[1]{\ensuremath{\statesymbol'^{(#1)}}}
\newcommand{\msg}{\boldsymbol{m}}

\newcommand{\reals}{\mathbb{R}}

\title{\ourtitle}

\author{%
  Miltiadis~Allamanis\\
  Microsoft Research\\
  Cambridge, UK \\
  \texttt{miallama@microsoft.com} \\
\And
  Marc~Brockschmidt\\
  Microsoft Research\\
  Cambridge, UK \\
  \texttt{mabrocks@microsoft.com}
\And
  Mahmoud~Khademi\thanks{Work done as an intern in Microsoft Research, Cambridge, UK.}\\
  Simon Fraser University\\
  Burnaby, BC, Canada\\
  \texttt{mkhademi@sfu.ca}
}

\begin{document}
\maketitle

\begin{abstract}
Learning tasks on source code (\ie, formal languages) have been considered
recently, but most work has tried to transfer natural language methods and does
not capitalize on the unique opportunities offered by code's known sematics.
For example, long-range dependencies induced by using the same variable or
function in distant locations are often not considered.
We propose to use graphs to represent both the syntactic and semantic structure
of code and use graph-based deep learning methods to learn to reason over
program structures.

In this work, we present how to construct graphs from source code and
how to scale Gated Graph Neural Networks training to such large graphs.
We evaluate our method on two tasks: \nametaskname, in which a network
attempts to predict the name of a variable given its usage, and \varmisusetask,
in which the network learns to reason about selecting the correct variable
that should be used at a given program location.
Our comparison to methods that use less structured program representations shows
the advantages of modeling known structure, and suggests that our models learn to
infer meaningful names and to solve the \varmisusetask task in many cases.
Additionally, our testing showed that \varmisusetask identifies a number
of bugs in mature open-source projects.
\end{abstract}

\section{Introduction}
The advent of large repositories of source code as well as scalable machine
learning methods naturally leads to the idea of ``big code'', \ie, largely
unsupervised methods that support software engineers by generalizing from
existing source code~\citep{allamanis2017survey}.
Currently, existing deep learning models of source code capture
its shallow, textual structure, \eg
 as a sequence of tokens~\citep{hindle2012naturalness,raychev2014code,allamanis2016convolutional},
 as parse trees~\citep{maddison2014structured,bielik2016phog}, or
 as a flat dependency networks of variables~\citep{raychev2015predicting}.
Such models miss out on the opportunity to capitalize on the rich and
well-defined semantics of source code.
In this work, we take a step to alleviate this by including two additional
signal sources in source code: data flow and type hierarchies.
We do this by encoding programs as graphs, in which edges represent syntactic
relationships (\eg ``token before/after'') as well as semantic relationships
(``variable last used/written here'', ``formal parameter for argument is called
\texttt{stream}'', \etc).
Our key insight is that exposing these semantics explicitly as structured input to a
machine learning model lessens the requirements on amounts of training data,
model capacity and training regime and allows us to solve tasks that are beyond
the current state of the art.

We explore two tasks to illustrate the advantages of exposing more
semantic structure of programs.
First, we consider the \nametaskname task~\citep{allamanis2014learning,raychev2015predicting}, in
which given some source code, the ``correct'' variable name is inferred
as a sequence of subtokens.
This requires some understanding of how a variable is used, \ie, requires
reasoning about lines of code far apart in the source file.
Secondly, we introduce the variable misuse prediction task
(\varmisusetask), in which the network aims to infer which variable should
be used in a program location.
To illustrate the task, \autoref{fig:runningExample} shows a slightly simplified
snippet of a bug our model detected in a popular open-source project.
Specifically, instead of the variable \id{clazz}, variable \id{first} should
have been used in the yellow highlighted slot.
Existing static analysis methods cannot detect such issues, even though a software
engineer would easily identify this as an error from experience.

To achieve high accuracy on these tasks, we need to learn representations of
program semantics.
For both tasks, we need to learn the \emph{semantic} role of a variable (\eg,
``is it a counter?'', ``is it a filename?'').
Additionally, for \varmisusetask, learning variable usage semantics (\eg, ``a
filename is needed here'') is required.
This ``fill the blank element'' task is related to methods for
learning distributed representations of natural language words, such as
Word2Vec~\citep{mikolov2013distributed} and GLoVe~\citep{pennington2014glove}.
However, we can learn from a much richer structure such as data flow
information.
This work is a step towards learning program representations, and we expect
them to be valuable in a wide range of other tasks, such as
 code completion (``this is the variable you are looking for'') and
 more advanced bug finding (``you should lock before using this object'').

\begin{figure}
  \begin{lstlisting}[frame=tlbr]
var clazz=classTypes["Root"].Single() as JsonCodeGenerator.ClassType;
Assert.NotNull(clazz);

var first=classTypes["RecClass"].Single() as JsonCodeGenerator.ClassType;
Assert.NotNull((*\placeholder{clazz}*));

Assert.Equal("string", first.Properties["Name"].Name);
Assert.False(clazz.Properties["Name"].IsArray);
   \end{lstlisting}%
   \vspace{-.35cm}%
   \caption{A snippet of a detected bug in RavenDB an open-source C\# project.
   The code has been slightly simplified. Our model detects correctly that the
   variable used in the highlighted (yellow) slot is incorrect. Instead, \id{first}
   should have been placed at the slot. We reported this problem which was fixed
   in \href{https://github.com/ravendb/ravendb/pull/4138}{PR 4138}.
   }\label{fig:runningExample}%
   \vspace{-.25cm}
\end{figure}

To summarize, our contributions are:
\begin{inparaenum}[(i)]
\item We define the \varmisusetask task as a challenge for machine
 learning modeling of source code, that requires to learn (some) semantics of
 programs (\cf \autoref{sec:task}).
\item We present deep learning models for solving the \nametaskname and \varmisusetask tasks by
  modeling the code's graph structure and learning program representations over
  those graphs (\cf \autoref{sec:models}).
\item We evaluate our models on a large dataset of 2.9 million lines of
 real-world source code, showing that our best model achieves 32.9\%
 accuracy on the \nametaskname task and 85.5\% accuracy on the \varmisusetask
 task, beating simpler baselines (\cf \autoref{sec:evaluation}).
\item We document practical relevance of \varmisusetask by summarizing some bugs that
 we found in mature open-source software projects (\cf
 \autoref{sec:bugs}).
\end{inparaenum}
Our implementation of graph neural networks (on a simpler task) can be found at
\url{https://github.com/Microsoft/gated-graph-neural-network-samples} and the
dataset can be found at \url{https://aka.ms/iclr18-prog-graphs-dataset}.


\section{Related Work}
Our work builds upon the recent field of using machine learning for source code artifacts~\citep{allamanis2017survey}.
For example,
 \citet{hindle2012naturalness,bhoopchand2016learning} model the code as a sequence of tokens,
while \citet{maddison2014structured,raychev2016probabilistic} model the syntax tree structure of code.
All works on language models of code find that predicting variable and method identifiers is one of 
biggest challenges in the task.

Closest to our work is the work of \citet{allamanis2015suggesting} who learn distributed representations
of variables using all their usages to predict their names.
However, they do not use data flow information and we are not aware of any model that does so.
\cite{raychev2015predicting} and \cite{bichsel2016statistical} use conditional random fields to model a
variety of relationships between variables, AST elements and types to predict variable names
and types (resp.\ to deobfuscate Android apps), but without considering the flow of data explicitly.
In these works, all variable usages are deterministically known beforehand (as the code
is complete and remains unmodified), as in \citet{allamanis2014learning,allamanis2015suggesting}.

Our work is remotely related to work on program synthesis using sketches \citep{solar2008program}
and automated code transplantation \citep{barr2015automated}.
However, these approaches require a set of specifications (\eg input-output examples, test suites) to complete
the gaps, rather than statistics learned from big code. These approaches can be thought as complementary to ours,
since we learn to statistically complete the gaps
without any need for specifications, by learning common variable usage
patterns from code.

Neural networks on graphs~\citep{gori2005new,li2015gated,defferrard2016convolutional,kipf2016semi,gilmer2017neural} adapt a variety of deep learning methods to graph-structured input. They have been
used in a series of applications, such as link prediction and classification~\citep{grover2016node2vec}
and semantic role labeling in NLP~\citep{marcheggiani2017encoding}. Somewhat related to source code
is the work of \citet{wang2017premise} who learn graph-based representations of mathematical formulas 
for premise selection in theorem proving.

\section{The \varmisusetask Task}
\label{sec:task}
Detecting variable misuses in code is a task that requires understanding and
reasoning about program semantics. To successfully tackle the task one needs
to infer the role and function of the program elements and understand how
they relate. For example,
given a program such as \rF{fig:runningExample}, the task is to automatically
detect that the marked use of \id{clazz} is a mistake and that \id{first}
should be used instead. While this task resembles standard code completion,
it differs significantly in its scope and purpose, by considering only
variable identifiers and a mostly complete program.

\paragraph{Task Description}
We view a source code file as a sequence of tokens $\tok_0 \dots \tok_N =
\tokenSeq$, in which some tokens $\tok_{\lambda_0}, \tok_{\lambda_1} \dots$ are
variables.
Furthermore, let $\varsInScope{\tok} \subset \varSet$ refer to the set of all
type-correct variables in scope at the location of $\tok$, \ie,
those variables that can be used at $\tok$ without raising a compiler error.
We call a token $tok_\lambda$ where we want to predict the correct variable
usage a \emph{slot}.
We define a separate task for each slot $\tok_{\lambda}$: Given $\tok_0 \ldots
\tok_{\lambda - 1}$ and $\tok_{\lambda + 1}, \ldots, \tok_{N}$, correctly
select $\tok_\lambda$ from $\varsInScope{\tok_\lambda}$.
For training and evaluation purposes, a correct solution is one that simply
matches the ground truth, but note that in practice, several possible
assignments could be considered correct (\ie, when several variables refer to
the same value in memory).


\section{Model: Programs as Graphs}
\label{sec:models}
In this section, we discuss how to transform program source code into \emph{program graphs}
and learn representations over them. These program graphs not only encode the program
text but also the semantic information that can be obtained using standard compiler tools.

\paragraph{Gated Graph Neural Networks}
Our work builds on Gated Graph Neural Networks~\citep{li2015gated} (GGNN) and we summarize
them here.
A graph $\graph = (\nodes, \edgelist, \nodefeats)$ is composed of a set of nodes
$\nodes$, node features $\nodefeats$, and a list of directed edge sets
$\edgelist = (\edges_1, \ldots, \edges_K)$ where $K$ is the number of edge
types.
We annotate each $\node \in \nodes$ with a real-valued vector $\nodefeat^{(\node)} \in
\reals^D$ representing the features of the node (\eg, the embedding of a string
label of that node).

We associate every node $\node$ with a state vector $\state{\node}$,
initialized from the node label $\nodefeat^{(\node)}$.
The sizes of the state vector and feature vector are typically the same, but
we can use larger state vectors through padding of node features.
To propagate information throughout the graph, ``messages'' of type $k$ are sent
from each $\node$ to its neighbors, where each message is computed from its
current state vector as $\msg_k^{(\node)} = f_{k}(\state{\node})$.
Here, $f_{k}$ can be an arbitrary function; we choose a linear layer in our case.
By computing messages for all graph edges at the same time, all states can be
updated at the same time.
In particular, a new state for a node $\node$ is computed by aggregating all
incoming messages as
$\tilde{\msg}^{(\node)} = g(\{ \msg_k^{(u)} \mid \text{there is an edge of type }
  k \text{ from } u \text{ to } v \})$.
$g$ is an aggregation function, which we implement as elementwise summation.
Given the aggregated message $\tilde{\msg}^{(\node)}$ and the current state
vector $\state{\node}$ of node $\node$, the state of the next time step
$\statePrime{\node}$ is computed as
$\statePrime{\node} = \textsc{GRU}(\tilde{\msg}^{(\node)}, \state{\node})$,
where $\textsc{GRU}$ is the recurrent cell function of gated recurrent unit
(GRU)~\citep{cho2014properties}.
The dynamics defined by the above equations are repeated for
a fixed number of time steps.
Then, we use the state vectors from the last time step as the node
representations.\footnote{Graph Convolutional Networks
  (GCN)~\citep{kipf2016semi,schlichtkrull2017modeling} would be a simpler
  replacement for GGNNs. They correspond to the special case of GGNNs in which
  no gated recurrent units are used for state updates and the number of
  propagation steps per GGNN layer is fixed to 1. Instead, several layers are
  used. In our experiments, GCNs generalized less well than GGNNs.}

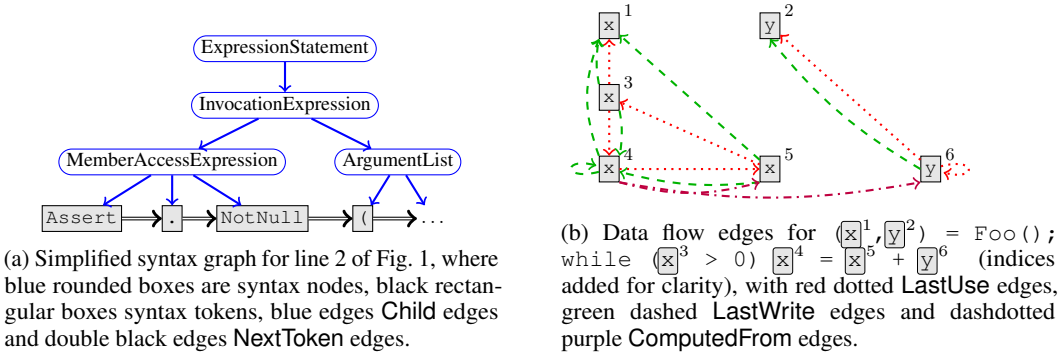
\begin{figure}
  \begin{subfigure}[b]{.47\textwidth}
    \begin{minipage}{\textwidth}
      \center
      \begin{tikzpicture}[codegraph]
         \node[SyntaxNode]
           (Expr) at (0,0)
           {ExpressionStatement};
         \node[SyntaxNode]
           (Invoc) at ($(Expr.south) + (0,-0.4)$)
           {InvocationExpression};
         \node[SyntaxNode]
           (Access) at ($(Invoc.south) + (-1.5,-0.4)$)
           {MemberAccessExpression};
         \node[SyntaxNode]
           (ArgList) at ($(Invoc.south) + (1.5,-0.4)$)
           {ArgumentList};
         \node[SyntaxToken]
           (Assert) at ($(Access.south) + (-1.2,-0.4)$)
           {\texttt{Assert}};
         \node[SyntaxToken]
           (Dot) at ($(Access.south) + (0,-0.4)$)
           {\texttt{.}};
         \node[SyntaxToken]
           (NotNull) at ($(Access.south) + (1.2,-0.4)$)
           {\texttt{NotNull}};
         \node[SyntaxToken]
           (OpenParen) at ($(ArgList.south west) + (0.2,-0.4)$)
           {\texttt{(}};
         \node[]
           (dotdotdot) at ($(ArgList.south east) + (-0.2,-0.4)$)
           {$\ldots$};
           
         \draw[->, thick, blue]
           (Expr) edge (Invoc)
           (Invoc) edge (Access)
           (Invoc) edge (ArgList)
           (Access) edge (Assert)
           (Access) edge (Dot)
           (Access) edge (NotNull)
           (ArgList) edge (OpenParen)
           (ArgList) edge (dotdotdot)
         ;

         \draw[->]
           (Assert) edge[double] (Dot)
           (Dot) edge[double] (NotNull)
           (NotNull) edge[double] (OpenParen)
           (OpenParen) edge[double] (dotdotdot)
         ;
       \end{tikzpicture}
    \end{minipage}
    \caption{\label{fig:SyntaxGraph}Simplified syntax graph for line 2 of
      \rF{fig:runningExample}, where blue rounded boxes are syntax nodes,
      black rectangular boxes syntax tokens, blue edges \textsf{Child} edges and
      double black edges \textsf{NextToken} edges.}
  \end{subfigure}
  \hspace*{0.05\textwidth}
  \begin{subfigure}[b]{.47\textwidth}
    \begin{minipage}{\textwidth}
      \begin{tikzpicture}[codegraph]
         \node[SyntaxToken, label=3:{\footnotesize $1$}]
           (x1) at (0,0)
           {x};
         \node[SyntaxToken, label=3:{\footnotesize $2$}]
           (y2) at ($(x1.north east) + (2,0)$)
           {y};
         \node[SyntaxToken, label=3:{\footnotesize $3$}]
           (x3) at ($(x1.south) + (0,-0.6)$)
           {x};
         \node[SyntaxToken, label=3:{\footnotesize $4$}]
           (x4) at ($(x3.south) + (0,-0.6)$)
           {x};
         \node[SyntaxToken, label=3:{\footnotesize $5$}]
           (x5) at ($(x4.north east) + (2,0)$)
           {x};
         \node[SyntaxToken, label=3:{\footnotesize $6$}]
           (y6) at ($(x5.north east) + (2,0)$)
           {y};

         \draw[thick, dotted, red]
           (x3) edge[->] (x1)
           (x3) edge[->] (x4)
           (x4) edge[->] (x5)
           (x5) edge[->] (x3)
           (y6) edge[->] (y2)
           (y6) edge[->, loop right] (y6)
         ;
         \draw[thick, dashed, darkgreen]
           (x3.north west) edge[->, bend left=10] (x1.south west)
           (x3.south east) edge[->, bend left=10] (x4.north east)
           (x4.north west) edge[->, bend left=20] (x1.south west)
           (x4) edge[->, loop left] (x4)
           (x5) edge[->] (x1)
           (x5.south west) edge[->, bend left=10] (x4)
           (y6.west) edge[->, bend left=10] (y2.south);
         ;
         \draw[thick, dashdotted, purple]
           (x4.south east) edge[->, bend right=15] (x5.south west)
           (x4.south east) edge[->, bend right=10] (y6.south west)
         ;
       \end{tikzpicture}
    \end{minipage}
    \caption{\label{fig:SemanticGraph}Data flow edges for
      \texttt{(\hlvar{x}{1},\hlvar{y}{2}) = Foo();
              while (\hlvar{x}{3} > 0)
                \hlvar{x}{4} = \hlvar{x}{5} + \hlvar{y}{6}}
      (indices added for clarity), with red dotted \textsf{LastUse} edges,
      green dashed \textsf{LastWrite} edges and dashdotted purple
      \textsf{ComputedFrom} edges.}
  \end{subfigure}%
  \vspace*{-.15cm}%
  \caption{\label{fig:ProgramGraph}Examples of graph edges used in program representation.}%
  \vspace{-.35cm}
\end{figure}

\paragraph{Program Graphs}
We represent program source code as graphs and use different edge types to model
syntactic and semantic relationships between different tokens.
The backbone of a program graph is the program's abstract syntax tree
(AST), consisting of \emph{syntax nodes} (corresponding to non-terminals in the
programming language's grammar) and \emph{syntax tokens} (corresponding to
terminals).
We label syntax nodes with the name of the nonterminal from the program's
grammar, whereas syntax tokens are labeled with the string that they represent.
We use \textsf{Child} edges to connect nodes according to the AST.
As this does not induce an order on children of a syntax node, we additionally
add \textsf{NextToken} edges connecting each syntax token to its successor.
An example of this is shown in \rF{fig:SyntaxGraph}.

To capture the flow of control and data through a program, we add additional edges
connecting different uses and updates of syntax tokens corresponding to
variables.
For such a token $\var$, let $\dfgReadPred{\var}$ be the set of syntax tokens
at which the variable could have been used last.
This set may contain several nodes (for example, when using a variable after a
conditional in which it was used in both branches), and even syntax tokens that
follow in the program code (in the case of loops).
Similarly, let $\dfgWritePred{\var}$ be the set of syntax tokens at which the
variable was last written to.
Using these, we add \textsf{LastRead} (resp. \textsf{LastWrite}) edges
connecting $\var$ to all elements of $\dfgReadPred{\var}$
(resp. $\dfgWritePred{\var}$).
Additionally, whenever we observe an assignment $\var = \mathit{expr}$, we
connect $\var$ to all variable tokens occurring in $\mathit{expr}$ using
\textsf{ComputedFrom} edges.
An example of such semantic edges is shown in \rF{fig:SemanticGraph}.

We extend the graph to chain all uses of the same variable using
\textsf{LastLexicalUse} edges (independent of data flow, \ie, in \texttt{if (\ldots) \{ \ldots\ \var \ldots \} else \{ \ldots\ \var \ldots \}}, we link the two
occurrences of \var).
We also connect \id{return} tokens to the method declaration using
\textsf{ReturnsTo} edges (this creates a ``shortcut'' to its name and type).
Inspired by \citet{rice2017detecting}, we connect arguments in method calls to
the formal parameters that they are matched to with \textsf{FormalArgName}
edges, \ie, if we observe a call \id{Foo(bar)} and a method declaration
\id{Foo(InputStream stream)}, we connect the \id{bar} token to the \id{stream}
token.
Finally, we connect every token corresponding to a variable to enclosing guard
expressions that use the variable with \textsf{GuardedBy} and
\textsf{GuardedByNegation} edges.
For example, in \texttt{if (x > y) \{ \ldots\ \underline{x} \ldots \} else \{
  \ldots\ \underline{y} \ldots \}},
we add a \textsf{GuardedBy} edge from \texttt{\underline{x}} (resp. a
\textsf{GuardedByNegation} edge from \texttt{\underline{y}}) to the AST node
corresponding to \texttt{x > y}.

Finally, for all types of edges we introduce their respective backwards edges (transposing the adjacency matrix),
doubling the number of edges and edge types. Backwards edges help with
propagating information faster across the GGNN and make the model more expressive.

\paragraph{Leveraging Variable Type Information}
We assume a statically typed language and that the source code can be compiled,
and thus each variable has a (known) type $\varType{\var}$.
To use it, we define a learnable embedding function $\typeEmbed{\tau}$ for known
types and additionally define an ``\textsc{UnkType}'' for all
unknown/unrepresented types.
We also leverage the rich type hierarchy that is available in many
object-oriented languages.
For this, we map a variable's type $\varType{\var}$ to the set of its
supertypes, \ie
$\allVarType{\var}=\{\tau : \varType{\var} \text{ implements type } \tau \} \cup \{ \varType{\var} \}$.
We then compute the type representation $\typeSetEmbed{\var}$ of a variable
$\var$ as the element-wise maximum of $\{ \typeEmbed{\tau} : \tau \in \allVarType{\var}\}$.
We chose the maximum here, as it is a natural pooling operation for representing
partial ordering relations (such as type lattices).
Using all types in $\allVarType{\var}$ allows us to generalize to unseen types
that implement common supertypes or interfaces.
For example, \id{List<K>} has multiple concrete types  (\eg \id{List<int>},
\id{List<string>}).
Nevertheless, these types implement a common interface (\id{IList}) and share
common characteristics.
During training, we randomly select a non-empty subset of $\allVarType{\var}$
which ensures training of all known types in the lattice. This acts both like
a dropout mechanism and allows us to learn a good representation for all types
in the type lattice.

\paragraph{Initial Node Representation} To compute the initial node state,
we combine information from the textual representation of the token and its
type. Concretely, we split the name of a node representing a token into subtokens (\eg \id{classTypes}
will be split into two subtokens \id{class} and \id{types}) on \id{camelCase} and
\id{pascal\_case}. We then average the embeddings of all subtokens to retrieve an
embedding for the node name. Finally, we concatenate the learned type
representation $\typeSetEmbed{\var}$, computed as discussed earlier, with the
node name representation, and pass it through a linear layer to obtain the initial
representations for each node in the graph.

\paragraph{Programs Graphs for \nametaskname}
Given a program and an existing variable $\var$, we build a program graph as
discussed above and then replace the variable name in all corresponding variable
tokens by a special \id{<SLOT>} token.
To predict a name, we use the initial node labels computed as the concatenation of
learnable token embeddings and type embeddings as discussed above, run GGNN
propagation for 8 time steps\footnote{We found fewer steps to be insufficient
  for good results and more propagation steps to not help substantially.\label{ft:8steps}}
and then compute a variable usage
representation by averaging the representations for all \id{<SLOT>} tokens.
This representation is then used as the initial state of a one-layer GRU,
which predicts the target name as a sequence
of subtokens (\eg, the name \texttt{inputStreamBuffer} is treated as the
sequence [\texttt{input}, \texttt{stream}, \texttt{buffer}]). We train this
graph2seq architecture using a maximum likelihood objective.
In \autoref{sec:evaluation}, we report the accuracy for
predicting the exact name and the F1 score for predicting its subtokens.

\paragraph{Program Graphs for \varmisusetask}
To model \varmisusetask with program graphs we need to modify the graph. First,
to compute a \emph{context representation} $\localContextRepr{\tok}$ for a slot
$\tok$ where we want to predict the used variable, we insert a 
new node $v_\id{<SLOT>}$ at
the position of $\tok$, corresponding to a ``hole'' at this point, and connect
it to the remaining graph using all applicable edges that do \emph{not} depend on the
chosen variable at the slot (\ie, everything but \textsf{LastUse},
\textsf{LastWrite}, \textsf{LastLexicalUse}, and \textsf{GuardedBy} edges).
Then, to compute the \emph{usage representation} $\usageRepr{\tok}{\var}$ of 
each candidate variable $\var$ at the target slot, we insert a ``candidate'' node
$v_{\tok,\var}$ for all $\var$ in $\varsInScope{\tok}$, and connect it to the
graph by inserting the \textsf{LastUse}, \textsf{LastWrite} and
\textsf{LastLexicalUse} edges that would be used if the variable were to be used
at this slot. Each of these candidate nodes represents the speculative placement
of the variable within the scope.

Using the initial node representations, concatenated with an extra bit
that is set to one for the candidate nodes $v_{\tok,\var}$, we run GGNN propagation
for 8 time steps.$^\text{\ref{ft:8steps}}$
The context and usage representation are then the final
node states of the nodes, \ie, $\localContextRepr{\tok} = \state{v_\id{<SLOT>}}$ and
$\usageRepr{\tok}{\var} = \state{v_{\tok, \var}}$.
Finally, the correct variable usage at the location is computed as $\argmax_{\var}
W [\localContextRepr{\tok}, \usageRepr{\tok}{\var}]$ where $W$ is a linear layer
that uses the concatenation of \localContextRepr{\tok} and \usageRepr{\tok}{\var}. We train using a max-margin objective.


\subsection{Implementation}
Using GGNNs for sets of large, diverse graphs requires some engineering effort,
as efficient batching is hard in the presence of diverse shapes.
An important observation is that large graphs are normally very sparse,
and thus a representation of edges as an adjacency list would usually be
advantageous to reduce memory consumption. 
In our case, this can be easily implemented using a sparse tensor representation,
allowing large batch sizes that exploit the parallelism of modern GPUs efficiently.
A second key insight is to represent a batch of graphs as one large graph with
many disconnected components.
This just requires appropriate pre-processing to make node identities unique.
As this makes batch construction somewhat CPU-intensive, we found it useful
to prepare minibatches on a separate thread.
Our TensorFlow~\citep{abadi2016tensorflow} implementation scales to 55 graphs per
second during training and 219 graphs per second during test-time using a single NVidia
GeForce GTX Titan X with graphs having on average 2,228 (median 936) nodes
and 8,350 (median 3,274) edges and 8 GGNN unrolling iterations, all 20 edge types (forward and backward edges for 10 original edge types) and the size of 
the hidden layer set to 64. The number of types of edges in the GGNN contributes
proportionally to the running time. For example, a GGNN run for our ablation
study using only the two most common edge types (\textsf{NextToken}, \textsf{Child}) achieves
105 graphs/second during training and 419 graphs/second at test time with the
same hyperparameters.
Our (generic) implementation of GGNNs is available at 
\url{https://github.com/Microsoft/gated-graph-neural-network-samples}, using a simpler
demonstration task.


\section{Evaluation}
\label{sec:evaluation}
\paragraph{Dataset}
We collected a dataset for the \varmisusetask task from open source \CSharp{}
projects on GitHub.
To select projects, we picked the top-starred (non-fork) projects in GitHub.
We then filtered out projects that we could not (easily) compile in full using
Roslyn\footnote{\url{http://roslyn.io}}, as we require a compilation to extract
precise type information for the code (including those types present in external
libraries).
Our final dataset contains 29 projects from a diverse set of domains (compilers,
databases, \ldots) with about 2.9 million non-empty lines of code.
A full table is shown in \autoref{appx:dataset}.

For the task of detecting variable misuses, we collect data from all
projects by selecting all variable usage locations, filtering out variable
declarations, where at least one other type-compatible replacement variable is
in scope.
The task is then to infer the correct variable that
originally existed in that location. Thus,
by construction there is at least one type-correct replacement variable, \ie
picking it would \emph{not} raise an error during type checking.
In our test datasets, at each slot there
are on average 3.8 type-correct alternative variables (median 3, $\sigma=2.6$).

\newcommand{\unseenTest}{\textsc{UnseenProjTest}\xspace}
\newcommand{\seenTest}{\textsc{SeenProjTest}\xspace}
From our dataset, we selected two projects as our development set.
From the rest of the projects, we selected three projects for \unseenTest to allow
testing on projects with completely unknown structure and types.
We split the remaining 23 projects into train/validation/test sets in the
proportion 60-10-30, splitting along files (\ie, \emph{all} examples from one
source file are in the same set).
We call the test set obtained like this \seenTest.

\paragraph{Baselines}
For \varmisusetask, we consider two bidirectional RNN-based baselines.
The local model (\localmodel) is a simple two-layer bidirectional GRU run over
the tokens before and after the target location.
For this baseline, \localContextRepr{\tok} is set to the slot representation
computed by the RNN, and the usage context of each variable
\usageRepr{\tok}{\var} is the embedding of the name and type of the
variable, computed in the same way as the initial node labels in the
\graphmodel.
This baseline allows us to evaluate how important the usage context information
is for this task.
The flat dataflow model (\avgbirnnmodel) is an extension to \localmodel,
where the usage representation \usageRepr{\tok}{\var} is computed using another
two-layer bidirectional RNN run over the tokens before/after each usage,
and then averaging over the computed representations at the variable token \var.
The local context, \localContextRepr{\tok}, is identical to
\localmodel.
\avgbirnnmodel is a significantly stronger baseline that already takes some
structural information into account, as the averaging over all variables usages
helps with long-range dependencies.
Both models pick the variable that maximizes
$\localContextRepr{\tok}^T \usageRepr{\tok}{\var}$.

For \nametaskname, we replace \localmodel by \avglblmodel, which uses a
log-bilinear model for 4 left and 4 right context tokens of each variable usage, and
then averages over these context representations (this corresponds to the model
in \citet{allamanis2015suggesting}).
We also test \avgbirnnmodel on \nametaskname, which essentially replaces the
log-bilinear context model by a bidirectional RNN.

\subsection{Quantitative Evaluation}

\begin{table}[t]
    \caption{Evaluation of models. \seenTest refers to the test set containing projects
      that have files in the training set, \unseenTest refers to projects that have no
      files in the training data. Results averaged over two runs.} \label{tbl:evaluation}
    \vspace{-3mm}
    \resizebox{\textwidth}{!}{%
    \scriptsize  
    \begin{tabular}{@{}lrrrrrrrrr@{}} \toprule
        & \multicolumn{4}{c}{\seenTest} && \multicolumn{4}{c}{\unseenTest} \\
        & \localmodel & \avglblmodel & \avgbirnnmodel & \graphmodel && \localmodel & \avglblmodel & \avgbirnnmodel & \graphmodel\\ \cmidrule{2-5} \cmidrule{7-10}
        \textbf{\varmisusetask} \\
        Accuracy (\%) & 50.0 & --- & 73.7 & \textbf{85.5} && 28.9 & --- & 60.2 & \textbf{78.2} \\
        PR AUC  & 0.788 & --- & 0.941 & \textbf{0.980} && 0.611 & --- & 0.895 & \textbf{0.958}\\
        \textbf{\nametaskname} \\
        Accuracy (\%) & --- & 36.1 & 42.9 & \textbf{53.6} && --- & 22.7 & 23.4 & \textbf{44.0} \\
        F1 (\%) & --- & 44.0 & 50.1 & \textbf{65.8} && --- & 30.6 & 32.0 & \textbf{62.0} \\
        \bottomrule
    \end{tabular}
    }
\end{table}

\begin{table}[t]
\caption{Ablation study for the GGNN model on \seenTest for the two
  tasks.} \label{tbl:ablation}
\vspace{-3mm}
\footnotesize \centering
\begin{tabular}{@{}lrr@{}} \toprule
  & \multicolumn{2}{c}{Accuracy (\%)} \\
Ablation Description & \varmisusetask & \nametaskname \\ \midrule
Standard Model (reported in \autoref{tbl:evaluation}) & 85.5 & 53.6\\
\midrule  
Only \textsf{NextToken}, \textsf{Child}, \textsf{LastUse}, \textsf{LastWrite} edges & 80.6 & 31.2 \\
Only semantic edges (all but \textsf{NextToken}, \textsf{Child}) & 78.4 & 52.9 \\
Only syntax edges (\textsf{NextToken}, \textsf{Child}) & 55.3 & 34.3 \\
\midrule  
Node Labels: Tokens instead of subtokens& 85.6 & 34.5 \\
Node Labels: Disabled& 84.3 & 31.8\\
\bottomrule
\end{tabular}
\end{table}

\autoref{tbl:evaluation} shows the evaluation results of the models for both
tasks.\footnote{\rSC{appx:perfCurves} additionally shows ROC and
  precision-recall curves for the \graphmodel model on the \varmisusetask task.}
As \localmodel captures very little information, it performs relatively badly.
\avglblmodel and \avgbirnnmodel, which capture information from many variable
usage sites, but do not explicitly encode the rich structure of the problem,
still lag behind the \graphmodel by a wide margin.
The performance difference is larger for \varmisusetask, since the structure and
the semantics of code are far more important within this setting.

\paragraph{Generalization to new projects} 
Generalizing across a diverse set of source code projects with different domains
is an important challenge in machine learning. We repeat the evaluation using
the \unseenTest set stemming from projects that have no files in the training set.
The right side of \autoref{tbl:evaluation} shows that our models
still achieve good performance, although it is slightly lower compared to \seenTest.
This is expected since the type lattice is mostly unknown in \unseenTest.

We believe that the dominant problem in applying a trained model to an unknown
project (\ie, domain) is the fact that its type hierarchy is unknown and the
used vocabulary (\eg in variables, method and class names, \etc) can differ
substantially.

\paragraph{Ablation Study}
To study the effect of some of the design choices for our models, we have run
some additional experiments and show their results in \autoref{tbl:ablation}.
First, we varied the edges used in the program graph.
We find that restricting the model to syntactic information has a large impact
on performance on both tasks, whereas restricting it to semantic edges seems to
mostly impact performance on \varmisusetask.
Similarly, the \textsf{ComputedFrom}, \textsf{FormalArgName} and
\textsf{ReturnsTo} edges give a small boost on \varmisusetask, but greatly
improve performance on \nametaskname.
As evidenced by the experiments with the node label representation, syntax node
and token names seem to matter little for \varmisusetask, but naturally have a
great impact on \nametaskname.

\subsection{Qualitative Evaluation}
\autoref{fig:suggestions} illustrates the predictions that \graphmodel makes on
a sample test snippet. The snippet recursively searches for the global
directives file by gradually descending into the root folder. Reasoning about the correct
variable usages is hard, even for humans, but the \graphmodel correctly predicts the
variable usages at all locations except two (slot 1 and 8). As a software engineer
is writing the code, it is imaginable that she may make a mistake misusing one
variable in the place of another. Since all variables are \id{string} variables,
no type errors will be raised. As the probabilities in \rF{fig:suggestions} suggest most potential
variable misuses can be flagged by the model yielding valuable warnings to
software engineers. Additional samples with comments
can be found in \autoref{appx:predictionsamples}.

\begin{figure}[t]
\include{figures/fullsuggestion}
\caption{\label{fig:suggestions}\varmisusetask predictions on slots within a snippet of the \seenTest set for the ServiceStack project. Additional
visualizations are available in \autoref{appx:predictionsamples}. The underlined tokens are the correct tokens.
The model has to select among a number of \id{string} variables at each slot, where all of them represent some
kind of path. The \graphmodel accurately predicts the correct variable usage in 11 out of the 13 slots reasoning
about the complex ways the variables interact among them.
}
\end{figure}

Furthermore, \autoref{appx:nnsamples} shows samples of pairs of code snippets that
share similar representations as computed
by the cosine similarity of the usage representation \usageRepr{\tok}{\var} of \graphmodel. The reader can notice
that the network learns to group variable usages that share semantic similarities together. For example,
checking for \id{null} before the use of a variable yields similar distributed representations across
code segments (Sample 1 in \autoref{appx:nnsamples}).

\subsection{Discovered Variable Misuse Bugs}
\label{sec:bugs}
We have used our \varmisusetask model to identify likely locations of bugs in
RavenDB (a document database) and Roslyn (Microsoft's \CSharp{} compiler
framework).
For this, we manually reviewed a sample of the top 500 locations in both
projects where our model was most confident about a choosing a variable
differing from the ground truth, and found three bugs in each of the projects.

\begin{figure}
    \begin{lstlisting}[frame=tlbr]
public ArraySegment<byte> ReadBytes(int length){
    int size = Math.Min(length, _len - _pos);
    var buffer = EnsureTempBuffer((*\placeholder{length}*));
    var used = Read(buffer, 0, size);
    \end{lstlisting}\vspace{-1em}
    \caption{A bug found (yellow) in RavenDB open-source project. The code 
    unnecessarily ensures that the buffer is of size \id{length}
    rather than \id{size} (which our model predicts as the
    correct variable here).}\label{fig:bug1}
\end{figure}

\begin{figure}
    \begin{lstlisting}[frame=tlbr]
  if (IsValidBackup(backupFilename) == false) {
    output("Error:"+(*\placeholder{backupLocation}*)+" doesn't look like a valid backup");
    throw new InvalidOperationException(
        (*\placeholder{backupLocation}*) + " doesn't look like a valid backup");
    \end{lstlisting}\vspace{-1em}
    \caption{A bug found (yellow) in the RavenDB open-source project. Although \id{backupFilename}
    is found to be invalid by \id{IsValidBackup}, the user is notified that \id{backupLocation} is invalid instead.}\label{fig:bug2}
\end{figure}

Figs.~\ref{fig:runningExample},\ref{fig:bug1},\ref{fig:bug2} show the issues
discovered in RavenDB.
The bug in \rF{fig:runningExample} was possibly caused by copy-pasting, and
cannot be easily caught by traditional methods.
A compiler will \emph{not} warn about unused variables (since \id{first} is
used) and virtually nobody would write a test testing another test.
\rF{fig:bug1} shows an issue that, although not critical,
can lead to increased memory consumption.
\rF{fig:bug2} shows another issue arising from a non-informative error
message.
We privately reported three additional bugs to the \href{http://roslyn.io}{Roslyn} developers, who have
fixed the issues in the meantime
(cf. \url{https://github.com/dotnet/roslyn/pull/23437}).
One of the reported bugs could cause a crash in Visual Studio when using certain
Roslyn features.

Finding these issues in widely released and tested code suggests that our model
can be useful during the software development process, complementing classic
program analysis tools.
For example, one usage scenario would be to guide the code reviewing process to
locations a \varmisusetask model has identified as unusual, or use it as a
prior to focus testing or expensive code analysis efforts.


\section{Discussion \& Conclusions}
Although source code is well understood and studied within other 
disciplines such as programming language research, it is a relatively
new domain for deep learning.
It presents novel opportunities compared to textual or perceptual data, as
its (local) semantics are well-defined and rich additional information can
be extracted using well-known, efficient program analyses.
On the other hand, integrating this wealth of structured information poses
an interesting challenge.
Our \varmisusetask task exposes these opportunities, going beyond simpler
tasks such as code completion.
We consider it as a first proxy for the core challenge of learning the
\emph{meaning} of source code, as it requires to probabilistically refine
standard information included in type systems.

\bibliographystyle{iclr2018_conference}
\bibliography{bibliography}

\newpage
\appendix
\section{Performance Curves}
\label{appx:perfCurves}
\begin{figure}
    \begin{subfigure}[b]{.47\textwidth}\centering
        \includegraphics[width=\textwidth]{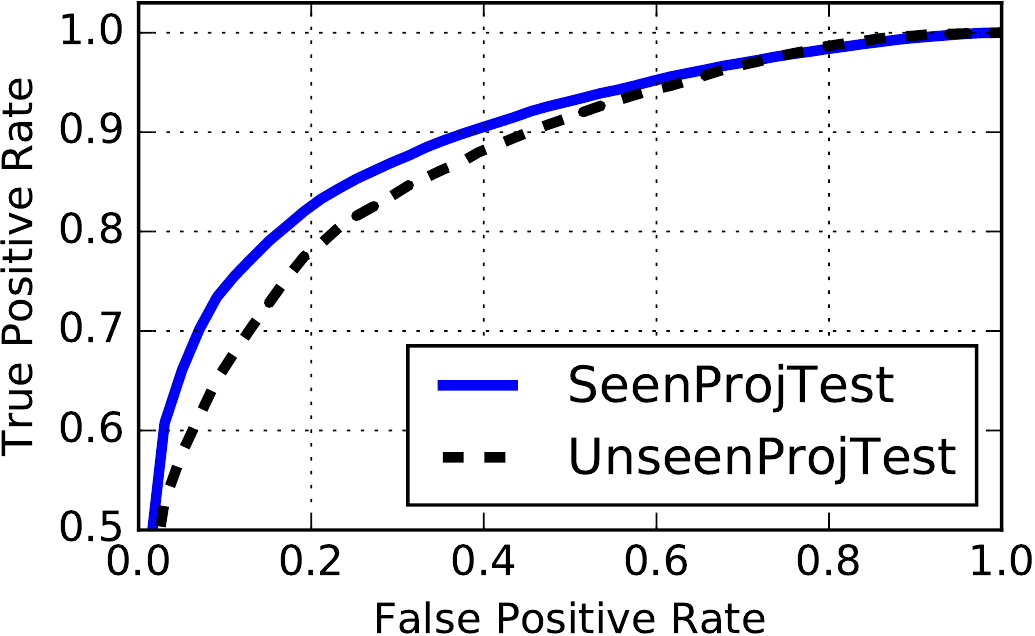}
        \caption{Precision-Recall Curve}
    \end{subfigure}
    \hspace*{0.05\textwidth}
    \begin{subfigure}[b]{.47\textwidth}\centering
        \includegraphics[width=\textwidth]{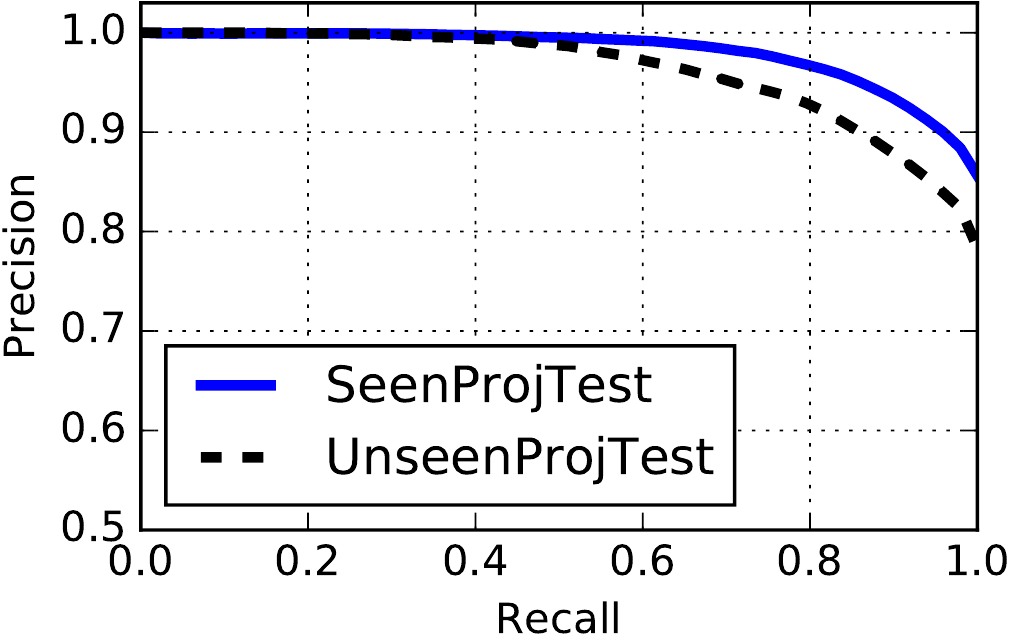}
        \caption{Receiver Operating Characteristic (ROC) Curve}
    \end{subfigure}
    \caption{Precision-Recall and ROC curves for the \graphmodel model on \varmisusetask. 
    Note that the $y$ axis starts from 50\%.}\label{fig:curves}
\end{figure}

\autoref{fig:curves} shows the ROC and precision-recall curves for the
\graphmodel model. As the reader may observe, setting a false positive rate to
10\% we get a true positive rate\footnote{A 10\% false positive rate is widely
accepted in industry, with 30\% as a maximum acceptable limit~\citep{bessey2010few}.}
of 73\% for the \seenTest and 69\% for
the unseen test. This suggests that this model can be practically used at a high
precision setting with acceptable performance.

\begin{table} \centering
    \caption{Performance of \graphmodel model on \varmisusetask per number of type-correct,
     in-scope candidate variables. Here we compute the performance of the full
     \graphmodel model
     that uses subtokens.}
    \begin{tabular}{@{}lrrrrrr@{}} \toprule
    \textbf{\# of candidates} & 2 & 3 & 4 & 5 & 6 or 7 & 8+ \\ \midrule
    \textbf{Accuracy on \seenTest} (\%) & 91.6 & 84.5 & 81.8 & 78.6 & 75.1 & 77.5\\ 
    \textbf{Accuracy on \unseenTest} (\%) & 85.7 & 77.1 & 75.7 & 69.0 & 71.5 & 62.4\\
    \bottomrule
    \end{tabular}
\end{table}

\section{\varmisusetask Prediction Samples}
\label{appx:predictionsamples}
Below we list a set of samples from our \seenTest projects with comments
about the model performance.
Code comments and formatting may have been
altered for typesetting reasons. The ground truth choice is underlined.

\begin{minipage}{\textwidth}
\textbf{Sample 1}
\begin{lstlisting}[frame=tlbr]
for (var port = (*\namedplaceholder{\#1}*); (*\namedplaceholder{\#2}*) < (*\namedplaceholder{\#3}*); (*\namedplaceholder{\#4}*)++)
{
  if (!activePorts.Contains((*\namedplaceholder{\#5}*)))
    return (*\namedplaceholder{\#6}*);
}
\end{lstlisting}
\namedplaceholder{\#1} \underline{\id{startingFrom}}: 97\%, \id{endingAt}: 3\%\\
\namedplaceholder{\#2} \underline{\id{port}}: 100\%, \id{startingFrom}: 0\%, \id{endingAt}: 0\%\\
\namedplaceholder{\#3} \underline{\id{endingAt}}: 100\%, \id{startingFrom}: 0\%, \id{port}: 0\%\\
\namedplaceholder{\#4} \underline{\id{port}}: 100\%, \id{startingFrom}: 0\%, \id{endingAt}: 0\%\\
\namedplaceholder{\#5} \underline{\id{port}}: 100\%, \id{startingFrom}: 0\%, \id{endingAt}: 0\%\\
\namedplaceholder{\#6} \underline{\id{port}}: 100\%, \id{startingFrom}: 0\%, \id{endingAt}: 0\%\\

\explain The model correctly predicts all variables in the loop.
\end{minipage}

\begin{minipage}{\textwidth}
\textbf{Sample 2}
\begin{lstlisting}[frame=tlbr]
var path = CreateFileName((*\namedplaceholder{\#1}*));
bitmap.Save((*\namedplaceholder{\#2}*), ImageFormat.Png);
return (*\namedplaceholder{\#3}*);
\end{lstlisting}
\namedplaceholder{\#1} \underline{\id{name}}: 86\%, \id{DIR\_PATH}: 14\%\\
\namedplaceholder{\#2} \underline{\id{path}}: 90\%, \id{name}: 8\%, \id{DIR\_PATH}: 2\%\\
\namedplaceholder{\#3} \underline{\id{path}}: 76\%, \id{name}: 16\%, \id{DIR\_PATH}: 8\%\\

\explain String variables are not confused their semantic role is inferred correctly.
\end{minipage}

\begin{minipage}{\textwidth}
\textbf{Sample 3}
\begin{lstlisting}[frame=tlbr]
[global::System.Diagnostics.DebuggerNonUserCodeAttribute]
public void MergeFrom(pb::CodedInputStream input) {
  uint tag;
  while ((tag = input.ReadTag()) != 0) {
    switch(tag) {
      default:
        input.SkipLastField();
        break;
      case 10: {
        (*\namedplaceholder{\#1}*).AddEntriesFrom(input, _repeated_payload_codec);
        break;
      }
    }
  }
}
\end{lstlisting}
\namedplaceholder{\#1} \id{Payload}: 66\%, \underline{\id{payload\_}}: 44\%\\

\explain The model is commonly confused by aliases, \ie variables that point to the
same location in memory. In this sample, either choice would have yielded identical
behavior.
\end{minipage}

\begin{minipage}{\textwidth}
\textbf{Sample 4}
\begin{lstlisting}[frame=tlbr]
public override bool IsDisposed
{
  get
  {
    lock ((*\namedplaceholder{\#1}*))
    {
       return (*\namedplaceholder{\#2}*);
    }
  }
}
\end{lstlisting}
\namedplaceholder{\#1} \underline{\id{\_gate}}: 99\%, \id{\_observers}: 1\%\\
\namedplaceholder{\#2} \underline{\id{\_isDisposed}}: 90\%, \id{\_isStopped}: 8\%, \id{HasObservers}: 2\%\\

\explain The \textsf{ReturnsTo} edge can help predict variables that otherwise would have been impossible.
\end{minipage}

\begin{minipage}{\textwidth}
\textbf{Sample 5}
\begin{lstlisting}[frame=tlbr]
/// <summary>
/// Notifies all subscribed observers about the exception.
/// </summary>
/// <param name="error">The exception to send to all observers.</param>
public override void OnError(Exception error)
{
    if ((*\namedplaceholder{\#1}*) == null)
        throw new ArgumentNullException(nameof((*\namedplaceholder{\#2}*)));

    var os = default(IObserver<T>[]);
    lock ((*\namedplaceholder{\#3}*))
    {
        CheckDisposed();

        if (!(*\namedplaceholder{\#4}*))
        {
            os = _observers.Data;
            _observers = ImmutableList<IObserver<T>>.Empty;
            (*\namedplaceholder{\#5}*) = true;
            (*\namedplaceholder{\#6}*) = (*\namedplaceholder{\#7}*);
        }
    }

    if (os != null)
    {
        foreach (var o in os)
        {
            o.OnError((*\namedplaceholder{\#8}*));
        }
    }
}
\end{lstlisting}
\namedplaceholder{\#1} \underline{\id{error}}: 93\%, \id{\_exception}: 7\%\\
\namedplaceholder{\#2} \underline{\id{error}}: 98\%, \id{\_exception}: 2\%\\
\namedplaceholder{\#3} \underline{\id{\_gate}}: 100\%, \id{\_observers}: 0\%\\
\namedplaceholder{\#4} \underline{\id{\_isStopped}}: 86\%, \id{\_isDisposed}: 13\%, \id{HasObservers}: 1\%\\
\namedplaceholder{\#5} \underline{\id{\_isStopped}}: 91\%, \id{\_isDisposed}: 9\%, \id{HasObservers}: 0\%\\
\namedplaceholder{\#6} \underline{\id{\_exception}}: 100\%, \id{error}: 0\%\\
\namedplaceholder{\#7} \underline{\id{error}}: 98\%, \id{\_exception}: 2\%\\
\namedplaceholder{\#8} \id{\_exception}: 99\%, \underline{\id{error}}: 1\%\\

\explain The model predicts the correct variables from all slots apart from the last.
Reasoning about the last one, requires interprocedural understanding of the code across
the class file.
\end{minipage}

\begin{minipage}{\textwidth}
\textbf{Sample 6}
\begin{lstlisting}[frame=tlbr]
private bool BecomingCommand(object message)
{
    if (ReceiveCommand((*\namedplaceholder{\#1}*)) return true;
    if ((*\namedplaceholder{\#2}*).ToString() == (*\namedplaceholder{\#3}*)) (*\namedplaceholder{\#4}*).Tell((*\namedplaceholder{\#5}*));
    else return false;
    return true;
}
\end{lstlisting}
\namedplaceholder{\#1} \underline{\id{message}}: 100\%, \id{Response}: 0\%, \id{Message}: 0\%\\
\namedplaceholder{\#2} \underline{\id{message}}: 100\%, \id{Response}: 0\%, \id{Message}: 0\%\\
\namedplaceholder{\#3} \id{Response}: 91\%, \underline{\id{Message}}: 9\%\\
\namedplaceholder{\#4} \underline{\id{Probe}}: 98\%, \id{AskedForDelete}: 2\%\\
\namedplaceholder{\#5} \underline{\id{Response}}: 98\%, \id{Message}: 2\%\\

\explain The model predicts correctly all usages except from the one in slot \#3. Reasoning
about this snippet requires additional semantic information about the intent of the code.
\end{minipage}

\begin{minipage}{\textwidth}
\textbf{Sample 7}
\begin{lstlisting}[frame=tlbr]
var response = ResultsFilter(typeof(TResponse), (*\namedplaceholder{\#1}*), (*\namedplaceholder{\#2}*), request);
\end{lstlisting}
\namedplaceholder{\#1} \underline{\id{httpMethod}}: 99\%, \id{absoluteUrl}: 1\%, \id{UserName}: 0\%, \id{UserAgent}: 0\%\\
\namedplaceholder{\#2} \underline{\id{absoluteUrl}}: 99\%, \id{httpMethod}: 1\%, \id{UserName}: 0\%, \id{UserAgent}: 0\%\\

\explain The model knows about selecting the correct string parameters because it matches them to the formal
parameter names.
\end{minipage}

\begin{minipage}{\textwidth}
\textbf{Sample 8}
\begin{lstlisting}[frame=tlbr]
 if ((*\namedplaceholder{\#1}*) >= (*\namedplaceholder{\#2}*))
    throw new InvalidOperationException(Strings_Core.FAILED_CLOCK_MONITORING);
\end{lstlisting}
\namedplaceholder{\#1} \underline{\id{n}}: 100\%, \id{MAXERROR}: 0\%, \id{SYNC\_MAXRETRIES}: 0\%\\
\namedplaceholder{\#2} \id{MAXERROR}: 62\%, \underline{\id{SYNC\_MAXRETRIES}}: 22\%, \id{n}: 16\%\\

\explain It is hard for the model to reason about conditionals, especially with rare constants
as in slot \#2.
\end{minipage}
\newpage
\section{Nearest Neighbor of \graphmodel Usage Representations}
\label{appx:nnsamples}
Here we show pairs of nearest neighbors based on the cosine similarity of 
the learned representations \usageRepr{\tok}{\var}. Each slot $\tok$ is marked
in dark blue and all usages of $\var$ are marked in yellow (\ie \placeholder{variableName}).
This is a set of hand-picked examples showing good
and bad examples. A brief description follows after each pair.

\begin{minipage}{\textwidth}
\textbf{Sample 1}
\begin{lstlisting}[frame=tlbr]
...
public void MoveShapeUp(BaseShape (*\placeholder{shape}*)) {
    if ((*\namedplaceholder{shape}*) != null) {
        for(int i=0; i < Shapes.Count -1; i++){
            if (Shapes[i] == (*\placeholder{shape}*)){
                Shapes.Move(i, ++i);
                return;
            }
        }
    }
}
...
\end{lstlisting}
\begin{lstlisting}[frame=tlbr]
...
lock(lockObject) {
    if ((*\namedplaceholder{unobservableExceptionHanler}*) != null)
        return false;
    (*\placeholder{unobservableExceptionHanler}*) = handler;
}
...
\end{lstlisting}
\explain Slots that are checked for null-ness have similar representations.
\end{minipage}

\begin{minipage}{\textwidth}
    \textbf{Sample 2}
    \begin{lstlisting}[frame=tlbr]
...
public IActorRef ResolveActorRef(ActorPath (*\placeholder{actorPath}*)){
  if(HasAddress((*\namedplaceholder{actorPath}*).Address))
    return _local.ResolveActorRef(RootGuardian, (*\placeholder{actorPath}*).ElementsWithUid);
  ...
...
    \end{lstlisting}
    \begin{lstlisting}[frame=tlbr]
...
ActorPath (*\placeholder{actorPath}*);
if (TryParseCachedPath(path, out actorPath)) {
    if (HasAddress((*\namedplaceholder{actorPath}*).Address)){
        if ((*\placeholder{actorPath}*).ToStringWithoutAddress().Equals("/"))
            return RootGuarding;
        ...
    }
    ...
}
...
    \end{lstlisting}
    \explain Slots that follow similar API protocols have similar representations.
    Note that the function \id{HasAddress} is a local function, seen only in the
    testset.
\end{minipage}

\begin{minipage}{\textwidth}
    \textbf{Sample 3}
    \begin{lstlisting}[frame=tlbr]
...
foreach(var (*\placeholder{filter}*) in configuration.Filters){
    GlobalJobFilter.Filters.Add((*\namedplaceholder{filter}*));
}
...
    \end{lstlisting}
    \begin{lstlisting}[frame=tlbr]
...
public void Count_ReturnsNumberOfElements(){
    _collection.Add((*\namedplaceholder{\_filterInstance}*));
    Assert.Equal(1, _collection.Count);
}
...
    \end{lstlisting}
    \explain Adding elements to a collection-like object yields
    similar representations.
\end{minipage}

\eat{
\section{Full Snippet Pasting Samples}
Below we present some of the suggestions when using
the full \varmisusetask structured prediction. The variables shown
at each placeholder correspond to the ground truth. Underlined
tokens represent UNK tokens. The top three allocations are shown
as well as the ground truth (if it is \emph{not} in the top 3 suggestions).
Red placeholders are the placeholders that need to be filled in when pasting.
All other placeholders are marked in superscript next to the relevant
variable.

\begin{minipage}{\textwidth}
\textbf{Sample 1}
\begin{lstlisting}[xleftmargin=0cm,frame=tlbr]
...
(*\hlplacehld{charsLeft}{1}*) = 0;
while ((*\hlplacehld{p}{2}*).IsRightOf((*\hlplacehld{selection}{3}*).Start))
{
    (*\hlplacehld{charsLeft}{4}*)++;
    (*\hlplacehld{p}{5}*).MoveUnit(_MOVEUNIT_ACTION.MOVEUNIT_PREVCHAR);
}
...
\end{lstlisting}
\begin{description}
\item[\hlreflarge{1}] \underline{\id{charsLeft}: 87\%}, \id{movesRight}: 8\%, \id{p}: 5\%
\item[\hlreflarge{2}] \underline{\id{p}: 96\%}, \id{selection}: 4\%, \id{bounds}: 1e-3
\item[\hlreflarge{3}] \underline{\id{selection}: 89\%}, \id{bounds}: 1\%, \id{p}: 8e-3
\item[\hlreflarge{4}] \id{movesRight}: 66\%, \underline{\id{charsLeft}: 16\%}, \id{p}: 1\%
\item[\hlreflarge{5}] \underline{\id{p}: 83\%}, \id{selection}: 11\%, \id{bounds}: 6\%
\end{description}
\end{minipage}

\begin{minipage}{\textwidth}
\textbf{Sample 2}
\begin{lstlisting}[xleftmargin=0cm,frame=tlbr]
...
HttpWebResponse response(*\hlref{0}*) = null;
XmlDocument xmlDocument(*\hlref{1}*) = new XmlDocument();
try
{
    using (Blog blog(*\hlref{3}*) = new Blog((*\hlplacehld{\_blogId}{4}*)))
        (*\hlplacehld{response}{5}*) = (*\hlplacehld{blog}{6}*).SendAuthenticatedHttpRequest((*\hlplacehld{notificationUrl}{7}*), 10000);

    // parse the results
    (*\hlplacehld{xmlDocument}{8}*).Load((*\hlplacehld{response}{9}*).GetResponseStream());
}
catch (Exception)
{
    throw;
}
finally
{
    if ((*\hlplacehld{response}{10}*) != null)
        (*\hlplacehld{response}{11}*).Close();
}
...
\end{lstlisting}
\begin{description}
\item[\hlreflarge{4}] \id{\_hostBlogId}: 12\%, \id{BlogId}: 10\%, \id{\_buttonId}: 10\%, \underline{\id{\_blogId}: 1\%}
\item[\hlreflarge{5}] \underline{\id{response}: 86\%}, \id{xmlDocument}: 5\%, \id{notificationUrl}: 3\%
\item[\hlreflarge{6}] \id{xmlDocument}: 84\%, \underline{\id{blog}: 12\%}, \id{response}: 2\%
\item[\hlreflarge{7}] \id{NotificationPollingTime}: 95\%, \id{CONTENT\_DISPLAY\_SIZE}: 2\%, \underline{\id{notificationUrl}: 1\%}
\item[\hlreflarge{8}] \underline{\id{xmlDocument}: 100\%}, \id{response}: 9e-4, \id{\_buttonDescription}: 4e-4 
\item[\hlreflarge{9}] \underline{\id{response}: 65\%}, \id{xmlDocument}: 30\%, \id{\_hostBlogId}: 4\%
\item[\hlreflarge{10}] \underline{\id{response}: 90\%}, \id{\_blogId}: 3\%, \id{CurrentImage}: 9e-3
\item[\hlreflarge{11}] \underline{\id{response}: 98\%}, \id{\_settingKey}: 1\%, \id{xmlDocument}: 9e-3
\end{description}
\end{minipage}

\begin{minipage}{\textwidth}
\textbf{Sample 3}
\begin{lstlisting}[xleftmargin=0cm,frame=tlbr]
...
protected override void Dispose(bool disposing(*\hlref{1}*))
{
    if ((*\hlplacehld{disposing}{2}*))
    {
        if ((*\hlplacehld{components}{3}*) != null)
            (*\hlplacehld{components}{4}*).Dispose();
    }
    base.Dispose((*\hlplacehld{disposing}{5}*));
}
...
\end{lstlisting}
\begin{description}
\item[\hlreflarge{2}] \underline{\id{disposing}: 100\%}, \id{commandIdentifier}: 4e-4, \id{components}: 1e-4
\item[\hlreflarge{3}] \underline{\id{components}: 100\%}, \id{disposing}: 3e-5, \id{commandIdentifier}: 2e-5
\item[\hlreflarge{4}] \underline{\id{components}: 100\%}, \id{disposing}: 9e-7, \id{CommandIdentifier}: 6e-9
\item[\hlreflarge{5}] \underline{\id{disposing}: 100\%}, \id{components}: 3e-5, \id{CommandIdentifier}: 2e-5
\end{description}
\end{minipage}

\begin{minipage}{\textwidth}
\textbf{Sample 4}
\begin{lstlisting}[xleftmargin=0cm,frame=tlbr]
...
tmpRange(*\hlref{1}*).Start.MoveAdjacentToElement(startStopParent(*\hlref{2}*),
                                         _ELEMENT_ADJACENCY.ELEM_ADJ_BeforeBegin);
if (tmpRange(*\hlref{3}*).IsEmptyOfContent())
{
    tmpRange(*\hlref{4}*).Start.MoveToPointer(selection(*\hlref{5}*).End);
    IHTMLElement endStopParent(*\hlref{6}*) = tmpRange(*\hlref{7}*).Start.GetParentElement(stopFilter(*\hlref{8}*));

    if ((*\hlplacehld{endStopParent}{9}*) != null 
            && (*\hlplacehld{startStopParent}{10}*).sourceIndex == (*\hlplacehld{endStopParent}{11}*).sourceIndex)
    {
        (*\hlplacehld{tmpRange}{12}*).Start
                .MoveAdjacentToElement((*\hlplacehld{endStopParent}{13}*),
                                             _ELEMENT_ADJACENCY.ELEM_ADJ_BeforeEnd);
        if ((*\hlplacehld{tmpRange}{14}*).IsEmptyOfContent())
        {
            (*\hlplacehld{tmpRange}{15}*).MoveToElement((*\hlplacehld{endStopParent}{16}*), true);
            if ((*\hlplacehld{maximumBounds}{17}*).InRange((*\hlplacehld{tmpRange}{18}*)) 
                                        && !((*\hlplacehld{endStopParent}{19}*) is IHTMLTableCell))
            {
                (*\hlplacehld{deleteParentBlock}{20}*) = true;
            }
        }
    }
}
...
\end{lstlisting}
\begin{description}
\item[\hlreflarge{9}] \id{startStopParent}: 97\%, \id{styleTagId}: 1\%, \id{tmpRange}: 1\%, \underline{\id{endStopParent}: 3e-3}
\item[\hlreflarge{10}] \underline{\id{startStopParent}: 100\%}, \id{tmpRange}: 2e-4, \id{maximumBounds}: 3e-5
\item[\hlreflarge{11}] \id{startStopParent}: 100\%, \id{styleTagId}: 2e-3, \underline{\id{endStopParent}: 1e-3}
\item[\hlreflarge{12}] \underline{\id{tmpRange}: 99\%}, \id{selection}: 9e-3, \id{startStopParent}: 2e-3
\item[\hlreflarge{13}] \id{startStopParent}: 96\%, \id{tmpRange}: 2\%, \underline{\id{endStopParent}: 1\%}
\item[\hlreflarge{14}] \underline{\id{tmpRange}: 98\%}, \id{selection}: 1\%, \id{maximumBounds}: 1\%
\item[\hlreflarge{15}] \underline{\id{tmpRange}: 98\%}, \id{selection}: 2\%, \id{maximumBounds}: 4e-3
\item[\hlreflarge{16}] \id{startStopParent}: 43\%, \id{styleTagId}: 29\%, \underline{\id{endStopParent}: 21\%}
\item[\hlreflarge{17}] \id{tmpRange}: 70\%, \id{selection}: 14\%, \underline{\id{maximumBounds}: 8\%}
\item[\hlreflarge{18}] \id{styleTagId}: 84\%, \underline{\id{tmpRange}: 5\%}, \id{selection}: 5\%
\item[\hlreflarge{19}] \id{startStopParent}: 98\%, \underline{\id{endStopParent}: 1\%}, \id{styleTagId}: 9e-3
\item[\hlreflarge{20}] \underline{\id{deleteParentBlock}: 90\%}, \id{startStopParent}: 4\%, \id{selection}: 3\%
\end{description}
\end{minipage}

\begin{minipage}{\textwidth}
\textbf{Sample 5}
\begin{lstlisting}[xleftmargin=0cm,frame=tlbr]
...
public static void GetImageFormat(string srcFileName(*\hlref{1}*), out string extension(*\hlref{2}*),
                                  out ImageFormat imageFormat(*\hlref{3}*))
{
    (*\hlplacehld{extension}{4}*) = Path.GetExtension((*\hlplacehld{srcFileName}{5}*))
                                            .ToLower(CultureInfo.InvariantCulture);
    if ((*\hlplacehld{extension}{6}*) == ".jpg" || (*\hlplacehld{extension}{7}*) == ".jpeg")
    {
        (*\hlplacehld{imageFormat}{8}*) = ImageFormat.Jpeg;
        (*\hlplacehld{extension}{9}*) = ".jpg";
    }
    else if ((*\hlplacehld{extension}{10}*) == ".gif")
    {
        (*\hlplacehld{imageFormat}{11}*) = ImageFormat.Gif;
    }
    else
    {
        (*\hlplacehld{imageFormat}{12}*) = ImageFormat.Png;
        (*\hlplacehld{extension}{13}*) = ".png";
    }
}
...
\end{lstlisting}
\begin{description}
\item[\hlreflarge{4}] \underline{\id{extension}: 64\%}, \id{imageFormat}: 36\%, \id{JPEG\_QUALITY}: 1e-4
\item[\hlreflarge{5}] \id{extension}: 98\%, \underline{\id{srcFileName}: 1\%}, \id{imageFormat}: 1e-3
\item[\hlreflarge{6}] \underline{\id{extension}: 97\%}, \id{imageFormat}: 1\%, \id{srcFileName}: 3e-4
\item[\hlreflarge{7}] \underline{\id{extension}: 75\%}, \id{JPG}: 4\%, \id{GIF}: 4\%
\item[\hlreflarge{8}] \underline{\id{imageFormat}: 100\%}, \id{extension}: 1e-5, \id{JPEG\_QUALITY}: 2e-6
\item[\hlreflarge{9}] \underline{\id{extension}: 93\%}, \id{imageFormat}: 2\%, \id{JPEG}: 9e-3
\item[\hlreflarge{10}] \underline{\id{extension}: 52\%}, \id{imageFormat}: 15\%, \id{ICO}: 6\%, \id{JPG}: 6\%, \id{GIF}: 6\%
\item[\hlreflarge{11}] \underline{\id{imageFormat}: 100\%}, \id{extension}: 4e-4, \id{JPEG\_QUALITY}: 1e-5
\item[\hlreflarge{12}] \underline{\id{imageFormat}: 99\%}, \id{JPEG\_QUALITY}: 4e-3, \id{extension}: 2e-3
\item[\hlreflarge{13}] \underline{\id{extension}: 66\%}, \id{JPG}: 6\%, \id{ICO}: 6\%, \id{GIF}: 6\%, \id{BMP}: 6\%
\end{description}
\end{minipage}

\begin{minipage}{\textwidth}
\textbf{Sample 6}
\begin{lstlisting}[xleftmargin=0cm,frame=tlbr]
...
BitmapData destBitmapData(*\hlref{1}*) = scaledBitmap(*\hlref{2}*).LockBits(
        new Rectangle(0, 0, destWidth(*\hlref{3}*), destHeight(*\hlref{4}*)),
        ImageLockMode.WriteOnly, scaledBitmap(*\hlref{5}*).PixelFormat);
try
{
    byte* s0(*\hlref{6}*) = (byte*)(*\hlplacehld{sourceBitmapData}{7}*).Scan0.ToPointer();
    int sourceStride(*\hlref{8}*) = (*\hlplacehld{sourceBitmapData}{9}*).Stride;
    byte* d0(*\hlref{10}*) = (byte*)(*\hlplacehld{destBitmapData}{11}*).Scan0.ToPointer();
    int destStride(*\hlref{12}*) = (*\hlplacehld{destBitmapData}{13}*).Stride;

    for (int y(*\hlref{14}*) = 0; y(*\hlref{15}*) < destHeight(*\hlref{16}*); y(*\hlref{17}*)++)
    {
        byte* d(*\hlref{18}*) = d0(*\hlref{19}*) + y(*\hlref{20}*) * destStride(*\hlref{21}*);
        byte* sRow(*\hlref{22}*) = s0(*\hlref{23}*) + ((int)(y(*\hlref{24}*) * yRatio(*\hlref{25}*)) 
                                + yOffset(*\hlref{26}*)) * sourceStride(*\hlref{27}*) + xOffset(*\hlref{28}*);
...
\end{lstlisting}
\begin{description}
\item[\hlreflarge{7}] \underline{\id{sourceBitmapData}: 72\%}, \id{destBitmapData}: 28\%, \id{bitmap}: 2e-6
\item[\hlreflarge{9}] \underline{\id{sourceBitmapData}: 90\%}, \id{destBitmapData}: 10\%, \id{bitmap}: 1e-4
\item[\hlreflarge{11}] \id{sourceBitmapData}: 75\%, \underline{\id{destBitmapData}: 25\%}, \id{s0}: 1e-5
\item[\hlreflarge{13}] \id{sourceBitmapData}: 83\%, \underline{\id{destBitmapData}: 17\%}, \id{bitmap}: 3e-4
\end{description}
\end{minipage}

\begin{minipage}{\textwidth}
\textbf{Sample 7}
\begin{lstlisting}[xleftmargin=0cm,frame=tlbr]
...
private static Stream GetStreamForUrl(string url(*\hlref{1}*), string pageUrl(*\hlref{2}*),
                                                             IHTMLElement element(*\hlref{3}*))
{
    if (UrlHelper.IsFileUrl(url(*\hlref{4}*)))
    {
        string path(*\hlref{5}*) = new Uri((*\hlplacehld{url}{6}*)).LocalPath;
        if (File.Exists((*\hlplacehld{path}{7}*)))
        {
            return File.OpenRead((*\hlplacehld{path}{8}*));
        }
        else
        {
            if (ApplicationDiagnostics.AutomationMode)
                Trace.WriteLine("File " + (*\hlplacehld{url}{9}*) + " not found");
            else
                Trace.Fail("File " + (*\hlplacehld{url}{10}*) + " not found");
            return null;
        }
    }
    else if (UrlHelper.IsUrlDownloadable(url(*\hlref{11}*)))
    {
        return HttpRequestHelper.SafeDownloadFile(url(*\hlref{12}*));
    }
    else
    {
...
\end{lstlisting}
\begin{description}
\item[\hlreflarge{6}] \underline{\id{url}: 96\%}, \id{element}: 2\%, \id{pageUrl}: 1\%
\item[\hlreflarge{7}] \underline{\id{path}: 86\%}, \id{url}: 14\%, \id{element}: 1e-3
\item[\hlreflarge{8}] \underline{\id{path}: 99\%}, \id{url}: 1\%, \id{pageUrl}: 4e-5
\item[\hlreflarge{9}] \id{path}: 97\%, \underline{\id{url}: 2\%}, \id{pagrUrl}: 4e-3\%
\item[\hlreflarge{10}] \id{path}: 67\%, \underline{\id{url}: 24\%}, \id{pageUrl}: 5\%
\end{description}
\end{minipage}

\begin{minipage}{\textwidth}
\textbf{Sample 8}
\begin{lstlisting}[xleftmargin=0cm,frame=tlbr]
...
public static void ApplyAlphaShift(Bitmap bitmap(*\hlref{1}*), double alphaPercentage(*\hlref{2}*))
{
    for (int y(*\hlref{3}*) = 0; y(*\hlref{4}*) < bitmap(*\hlref{5}*).Height; y(*\hlref{6}*)++)
    {
        for (int x(*\hlref{7}*) = 0; x(*\hlref{8}*) < bitmap(*\hlref{9}*).Width; x(*\hlref{10}*)++)
        {
            Color c(*\hlref{11}*) = bitmap.GetPixel(x(*\hlref{12}*), y(*\hlref{13}*));
            if ((*\hlplacehld{c}{14}*).(*\underline{\id{A}}*) > 0) //never make transparent pixels non-transparent
            {
                int newAlphaValue(*\hlref{15}*) = (int)((*\hlplacehld{c}{16}*).(*\underline{\id{A}}*) * (*\hlplacehld{alphaPercentage}{17}*));
                //value must be between 0 and 255
                (*\hlplacehld{newAlphaValue}{18}*) = Math.Max(0, Math.Min(255, (*\hlplacehld{newAlphaValue}{19}*)));
                (*\hlplacehld{bitmap}{20}*).SetPixel((*\hlplacehld{x}{21}*), (*\hlplacehld{y}{22}*), Color.FromArgb((*\hlplacehld{newAlphaValue}{23}*), (*\hlplacehld{c}{24}*)));
            }
            else
                (*\hlplacehld{bitmap}{25}*).SetPixel((*\hlplacehld{x}{26}*), (*\hlplacehld{y}{27}*), (*\hlplacehld{c}{28}*));
        }
    }
}
...
\end{lstlisting}
\begin{description}
\item[\hlreflarge{14}] \id{alphaPercentage}: 52\%, \id{bitmap}: 32\%, \underline{\id{c}: 13\%}
\item[\hlreflarge{16}] \id{bitmap}: 67\%, \id{alphaPercentage}: 27\%, \underline{\id{c}: 4\%}
\item[\hlreflarge{17}] \underline{\id{alphaPercentage}: 85\%}, \id{c}: 6\%, \id{JPEQ\_QUALITY}: 3\%
\item[\hlreflarge{18}] \underline{\id{newAlphaValue}: 51\%}, \id{bitmap}: 24\%, \id{alphaPercentage}: 11\%
\item[\hlreflarge{19}] \underline{\id{newAlphaValue}: 86\%}, \id{y}: 4\%, \id{alphaPercentage}: 3\%
\item[\hlreflarge{20}] \underline{\id{bitmap}: 100\%}, \id{c}: 4e-3, \id{JPEG\_QUALITY}: 3e-4
\item[\hlreflarge{21}] \id{bitmap}: 98\%, \underline{\id{x}: 9e-2}, \id{c}: 8e-3
\item[\hlreflarge{22}] \id{c}: 50\%, \id{bitmap}: 49\%, \id{newAlphaValue}: 2e-3, \underline{\id{y}: 3e-8}
\item[\hlreflarge{23}] \id{alphaPercentage}: 42\%, \id{JPEG\_QUALITY}: 40\%, \id{bitmap}: 10\%, \underline{\id{newAlphaValue}: 3\%}
\item[\hlreflarge{24}] \id{newAlphaValue}: 60\%, \id{alphaPercentage}: 25\%, \underline{\id{c}: 5\%}
\item[\hlreflarge{25}] \underline{\id{bitmap}: 100\%}, \id{c}: 8e-4, \id{alphaPercentage}: 3e-4
\item[\hlreflarge{26}] \id{bitmap}: 88\%, \underline{\id{x}: 9\%}, \id{c}: 2\%
\item[\hlreflarge{27}] \id{c}: 79\%, \id{bitmap}: 18\%, \id{JPEG\_QUALITY}: 1\%, \underline{\id{y}: 3e-3}
\item[\hlreflarge{28}] \underline{\id{c}: 82\%}, \id{y}: 6\%, \id{x}: 4\%
\end{description}
\end{minipage}

\begin{minipage}{\textwidth}
\textbf{Sample 9}
\begin{lstlisting}[xleftmargin=0cm,frame=tlbr]
...
string s(*\hlref{1}*) = (string)(*\hlplacehld{Value}{2}*);
byte[] data(*\hlref{3}*);

Guid g(*\hlref{4}*);
if ((*\hlplacehld{s}{5}*).Length == 0)
{
    (*\hlplacehld{data}{6}*) = CollectionUtils.ArrayEmpty<byte>();
}
else if (ConvertUtils.TryConvertGuid((*\hlplacehld{s}{7}*), out (*\hlplacehld{g}{8}*)))
{
    (*\hlplacehld{data}{9}*) = (*\hlplacehld{g}{10}*).ToByteArray();
}
else
{
    (*\hlplacehld{data}{11}*) = Convert.FromBase64String((*\hlplacehld{s}{12}*));
}

SetToken(JsonToken.Bytes, data(*\hlref{13}*), false);
return data(*\hlref{15}*);
...
\end{lstlisting}
\begin{description}
\item[\hlreflarge{2}] \id{t}: 58\%, \underline{\id{Value}: 12\%}, \id{\_tokenType}: \%
\item[\hlreflarge{5}] \id{TokenType}: 44\%, \id{QuoteChar}: 43\%, \id{\_currentPosition}: 4\%, \underline{\id{s}: 9e-3}
\item[\hlreflarge{6}] \id{\_tokenType}: 74\%, \underline{\id{data}: 20\%}, \id{\_currentState}: 5\%
\item[\hlreflarge{7}] \id{QuoteChar}: 31\%, \id{ValueType}: 26\%, \id{Path}: 9\%, \underline{\id{s}: 3e-4}
\item[\hlreflarge{8}] \underline{\id{g}: 100\%}, \id{data}: 6e-5, \id{t}: 3e-5
\item[\hlreflarge{9}] \underline{\id{data}: 99\%}, \id{\_tokenType}: 5e-3, \id{ValueType}: 9e-4
\item[\hlreflarge{10}] \underline{\id{g}: 99\%}, \id{data}: 6e-3, \id{\_currentState}: 2e-3
\item[\hlreflarge{11}] \underline{\id{data}: 66\%}, \id{\_tokenType}: 31\%, \id{\_currentState}: 6e-3
\item[\hlreflarge{12}] \underline{\id{s}: 74\%}, \id{Value}: 20\%, \id{t}: 3\%
\end{description}
\end{minipage}

% ...

% ...
}

\section{Dataset}
\label{appx:dataset}
The collected dataset and its characteristics are listed in \autoref{tbl:dataset}.
The full dataset as a set of projects and its parsed JSON will become available online.

\newcommand{\dev}{$^{Dev}\xspace$}
\newcommand{\testonly}{$^\dag\xspace$}

\begin{table}[hp]
\caption{Projects in our dataset. Ordered alphabetically. kLOC measures the number of
non-empty lines of C\# code. Projects marked with \dev were used as a
development set. Projects marked with \testonly were in the test-only dataset.
The rest of the projects were split into train-validation-test. The
dataset contains in total about 2.9MLOC.}\label{tbl:dataset}
\resizebox{\textwidth}{!}{
\begin{tabular}{lrrrrp{5.3cm}} \toprule
Name & Git SHA & kLOCs & Slots & Vars & Description \\ \midrule
Akka.NET& \id{719335a1} & 240 & 51.3k& 51.2k& Actor-based Concurrent \& Distributed Framework\\
AutoMapper& \id{2ca7c2b5} & 46& 3.7k& 10.7k & Object-to-Object Mapping Library \\
BenchmarkDotNet& \id{1670ca34} & 28& 5.1k & 6.1k& Benchmarking Library\\
BotBuilder& \id{190117c3} & 44 & 6.4k & 8.7k& SDK for Building Bots\\
choco& \id{93985688} & 36 & 3.8k& 5.2k& Windows Package Manager\\
commandline\testonly& \id{09677b16} & 11 & 1.1k& 2.3k& Command Line Parser \\
CommonMark.NET\dev& \id{f3d54530} & 14& 2.6k& 1.4k& Markdown Parser\\
Dapper& \id{931c700d} & 18& 3.3k& 4.7k& Object Mapper Library\\
EntityFramework& \id{fa0b7ec8}  & 263& 33.4k& 39.3k& Object-Relational Mapper\\
Hangfire& \id{ffc4912f} & 33& 3.6k& 6.1k& Background Job Processing Library\\
Humanizer\testonly & \id{cc11a77e} & 27& 2.4k & 4.4k& String Manipulation and Formatting\\
Lean\testonly & \id{f574bfd7} & 190& 26.4k& 28.3k& Algorithmic Trading Engine\\
Nancy& \id{72e1f614} & 70 & 7.5k& 15.7& HTTP Service Framework \\
Newtonsoft.Json& \id{6057d9b8} & 123& 14.9k& 16.1k& JSON Library\\
Ninject& \id{7006297f} & 13& 0.7k& 2.1k& Code Injection Library\\
NLog& \id{643e326a}  & 75 & 8.3k& 11.0k& Logging Library\\
Opserver& \id{51b032e7} & 24& 3.7k& 4.5k& Monitoring System\\
OptiKey& \id{7d35c718} & 34& 6.1k& 3.9k& Assistive On-Screen Keyboard \\
orleans& \id{e0d6a150} & 300& 30.7k& 35.6k& Distributed Virtual Actor Model \\
Polly& \id{0afdbc32} & 32& 3.8k& 9.1k& Resilience \& Transient Fault Handling Library \\
quartznet& \id{b33e6f86} & 49 & 9.6k & 9.8k & Scheduler \\
ravendb\dev& \id{55230922} & 647& 78.0k& 82.7k& Document Database \\
RestSharp& \id{70de357b} & 20& 4.0k& 4.5k & REST and HTTP API Client Library \\
Rx.NET& \id{2d146fe5} & 180 & 14.0k& 21.9k& Reactive Language Extensions \\
scriptcs&\id{f3cc8bcb} & 18& 2.7k& 4.3k& C\# Text Editor\\
ServiceStack& \id{6d59da75} & 231& 38.0k& 46.2k & Web Framework\\
ShareX& \id{718dd711} & 125 & 22.3k& 18.1k& Sharing Application\\
SignalR& \id{fa88089e} & 53& 6.5k& 10.5k& Push Notification Framework\\
Wox& \id{cdaf6272} & 13& 2.0k& 2.1k& Application Launcher\\ \bottomrule
\end{tabular}
}
\end{table}

For this work, we released a large portion of the data, with the exception of
projects with a GPL license. The data can be found at \url{https://aka.ms/iclr18-prog-graphs-dataset}.
Since we are excluding some projects from the data, below we report the results, averaged
over three runs, on the published dataset:

\begin{tabular}{lrr} \toprule
    & Accuracy (\%) & PR AUC \\ \midrule



\seenTest   & 84.0 & 0.976 \\
\unseenTest & 74.1 & 0.934\\ \bottomrule
\end{tabular}

\end{document}